\documentclass[10pt,twocolumn,letterpaper]{article}

\usepackage{iccv}              

\usepackage{comment}

\usepackage{booktabs} 
\usepackage{siunitx}  
\usepackage{multirow} 

\usepackage[most]{tcolorbox}

\newtcolorbox{disclaimerbox}{
    colback=gray!10,     
    colframe=gray!40,    
    boxrule=0.5pt,       
    arc=4pt,             
    auto outer arc,
    boxsep=5pt,
    left=6pt,
    right=6pt,
    top=4pt,
    bottom=4pt,
    enhanced jigsaw
}

\definecolor{iccvblue}{rgb}{0.21,0.49,0.74}
\usepackage[pagebackref,breaklinks,colorlinks,allcolors=iccvblue]{hyperref}

\def\paperID{*****} 
\def\confName{ICCV}
\def\confYear{2025}

\title{AIM 2025 Challenge on Real-World RAW Image Denoising}

\author{
Feiran Li$^\diamond$\thanks{First two authors contribute equally.\\
M. Conde (\href{mailto:marcos.conde@uni-wuerzburg.de}{marcos.conde@uni-wuerzburg.de}) is the corresponding author. \\
AIM 2025 webpage: \href{https://cvlai.net/aim/2025}{https://cvlai.net/aim/2025}\\
Challenge and Code: \href{https://www.codabench.org/competitions/8529/}{Codalab Competition} and \href{https://sonyresearch.github.io/AIM2025_Denoise_Challenge/}{Challenge Site} } \quad
Jiacheng Li$^\diamond$ \quad
Marcos V. Conde$^{\dagger\ddagger}$ \quad
Beril Besbinar$^\diamond$ \quad
Vlad Hosu$^\diamond$ \\
Daisuke Iso$^\diamond$ \quad
Radu Timofte$^\dagger$ \\ \\
{$^\diamond$ Sony Research \quad $^\dagger$ University of W\"urzburg, Computer Vision Lab}
}

\begin{document}
\maketitle
\begin{abstract}
We introduce the AIM 2025 Real-World RAW Image Denoising Challenge, aiming to advance efficient and effective denoising techniques grounded in data synthesis. The competition is built upon a newly established evaluation benchmark featuring challenging low-light noisy images captured in the wild using five different DSLR cameras. Participants are tasked with developing novel noise synthesis pipelines, network architectures, and training methodologies to achieve high performance across different camera models. Winners are determined based on a combination of performance metrics, including full-reference measures (PSNR, SSIM, LPIPS), and non-reference ones (ARNIQA, TOPIQ). By pushing the boundaries of camera-agnostic low-light RAW image denoising trained on synthetic data, the competition promotes the development of robust and practical models aligned with the rapid progress in digital photography. We expect the competition outcomes to influence multiple domains, from  image restoration to night-time autonomous driving.

\end{abstract}    
\section{Introduction}
\label{sec:intro}

The pursuit of high-fidelity digital imaging under adverse lighting conditions remains a formidable and critical challenge in computational photography. Low-light scenarios inherently force a trade-off between noise and signal, leading to images where crucial details are obscured by sensor artifacts. While processing RAW image data offers the most potential for faithful restoration by bypassing in-camera processing pipelines~\cite{abdelhamed2019noise, conde2025ntire, conde2024toward}, it also exposes the complex, device-specific nature of noise. One of the major bottlenecks hindering progress is the reliance on extensive, paired datasets to obtain robust denoising models for a specific camera. This dependency makes it impractical to develop solutions that can generalize effectively across the vast and ever-growing ecosystem of digital cameras.

To address this critical gap and catalyze innovation, we introduce the AIM 2025 Real-World RAW Image Denoising Challenge. This competition is fundamentally designed to push a step further from camera-specific methods and towards the development of universal, camera-agnostic denoising solutions grounded in advanced data synthesis. The challenge tasks participants with creating novel noise modeling pipelines and learning-based architectures that are not only perform well on real-world scenes, but also generalize to various cameras.

To facilitate this, we have established a new, challenging evaluation benchmark comprising low-light RAW images captured with five distinct DSLR camera models. To better align with real-world scenarios, we consider both in-door paired scenes and out-door in-the-wild scenes. Consequently, the performance of submissions is assessed through a comprehensive suite of metrics, combining established full-reference measures like PSNR and SSIM with modern perceptual (e.g., LPIPS~\cite{zhang2018perceptual}) and non-reference (e.g., ARNIQA~\cite{agnolucci2024arniqa}, TOPIQ~\cite{chen2024topiq}) evaluations to provide a holistic view of image quality.

By pushing the boundaries of camera-agnostic RAW image denoising, this challenge aims to foster the development of practical, high-performance models that align with the rapid pace of innovation in digital imaging. We anticipate that the proposed benchmark and outcomes of this competition will inspire new methodologies, not only advancing the state of the art in academic research but also influencing real-world applications ranging from consumer night photography to the safety-critical domain of autonomous driving.

\begin{figure*}[t]
    \centering
    \includegraphics[width=0.9\linewidth]{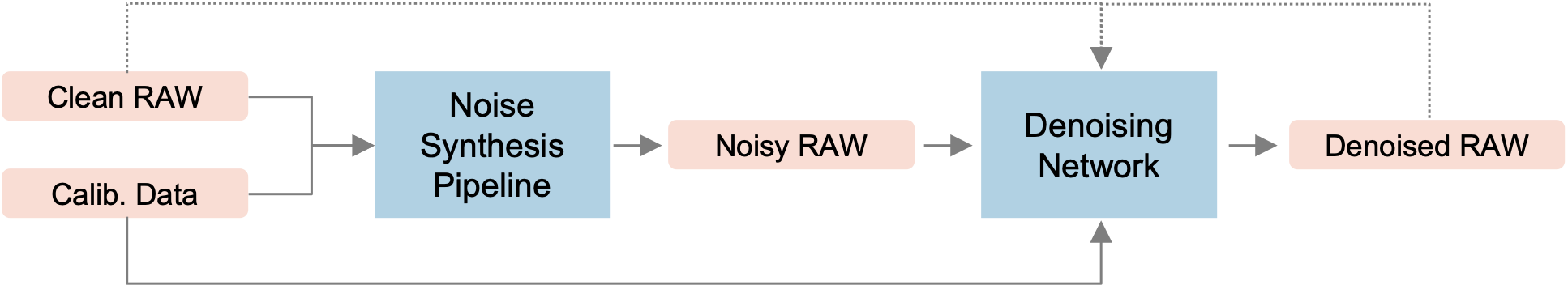}
    \caption{Illustration of a classical RAW Image Denoising Pipeline~\cite{li2025noise}.}
    \label{fig:rawdenoise}
\end{figure*}

\begin{figure*}[t]
    \centering
    \includegraphics[width=0.9\linewidth]{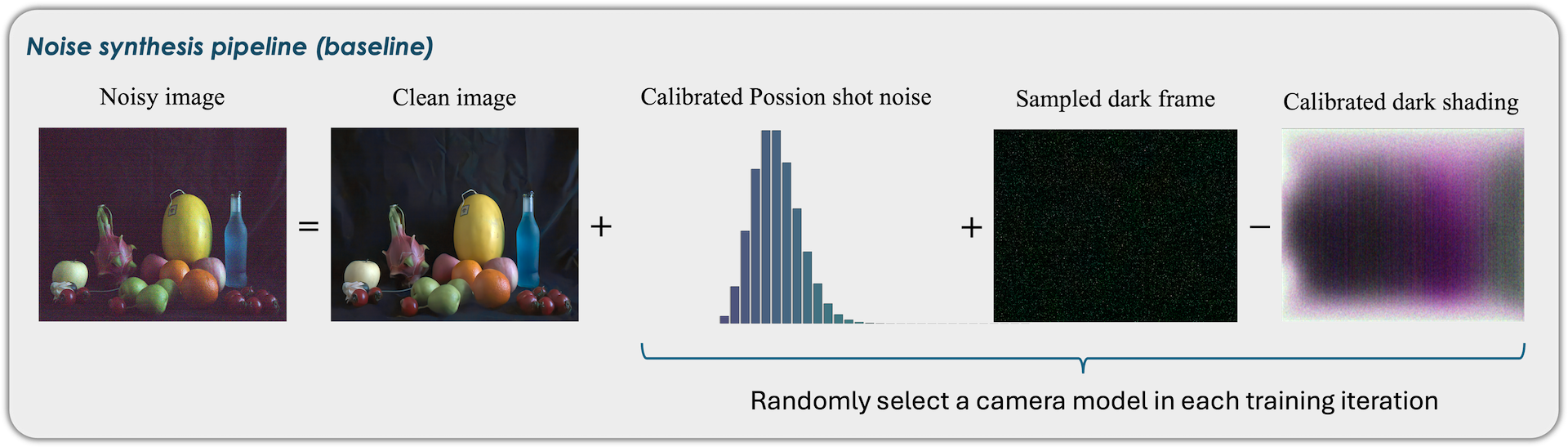}
    \caption{Illustration of the baseline model for noise synthesis~\cite{li2025noise}.}
    \label{fig:noise}
\end{figure*}


\vspace{-2mm}
\paragraph{Related Challenges}
This challenge is one of the AIM 2025~\footnote{\url{https://www.cvlai.net/aim/2025/}} workshop associated challenges on: high FPS non-uniform motion deblurring~\cite{aim2025highfps}, rip current segmentation~\cite{aim2025ripseg}, inverse tone mapping~\cite{aim2025tone}, robust offline video super-resolution~\cite{aim2025videoSR}, low-light raw video denoising~\cite{aim2025videodenoising}, screen-content video quality assessment~\cite{aim2025scvqa}, 
perceptual image super-resolution~\cite{aim2025perceptual}, 
efficient real-world deblurring~\cite{aim2025efficientdeblurring}, 4K super-resolution on mobile NPUs~\cite{aim20254ksr}, efficient denoising on smartphone GPUs~\cite{aim2025efficientdenoising}, efficient learned ISP on mobile GPUs~\cite{aim2025efficientISP}, and stable diffusion for on-device inference~\cite{aim2025sd}. Descriptions of the datasets, methods, and results can be found in the corresponding challenge reports.

\section{Related Work}
\label{sec:related_work}

Data synthesis offers a promising solution to the problem of limited training data. In the context of image denoising, it involves constructing noise models and applying them to clean images to generate synthetic noisy-clean pairs.

\subsection{Camera-specific RAW image denoising}
Noise synthesis and denoising network training are typically conducted in a camera-specific manner to ensure accurate modeling of the noise characteristics. For example, ELD~\cite{wei2021physics} decompose the overall noise profile to isolated components and proposes modeling them statistically. Monakhova~\etal~\cite{monakhova2022dancing} employ generative adversarial network for data synthesis for starlight video denoising. Cao~\etal~\cite{cao2023physics} introduce a normalizing flow framework to connect noise components to camera ISO. Feng~\etal~\cite{feng2023physics} propose a deep proxy network for profiling the i.i.d components of signal-independent noise. There are also efforts to synthesize noise without explicit parametric modeling. For example, Zhang~\etal~\cite{zhang2021rethinking} directly sample dark frames from the sensors to represent signal-independent noise. Mosleh~\etal~\cite{Mosleh2024nonparam} propose a histogram-based methods for non-parametric noise modeling. Li~\etal~\cite{li2025noise} demonstrate that certain noise calibration procedures can be simplified to reduce effort without compromising denoising performance.

\subsection{Camera-agnostic RAW image denoising}
Camera-agnostic denoising has attracted increasing attention due to its greater flexibility in real-world applications. For example, LED~\cite{jin2023lighting} presents a sensor-agnostic pre-training and finetuning framework based on the noise model developed in ELD~\cite{wei2021physics}. Zou~\etal~\cite{zou2025calibration} integrate a fine-grained statistical noise model and contrastive learning strategy to estimate noise parameters on the inputs. Feng~\etal~\cite{feng2025yond} propose coarse-to-fine noise estimation  and expectation matched variance-stabilizing transform to identify noise characteristics and remove its camera dependency.

\begin{figure*}[t]
    \centering
    \includegraphics[width=\linewidth]{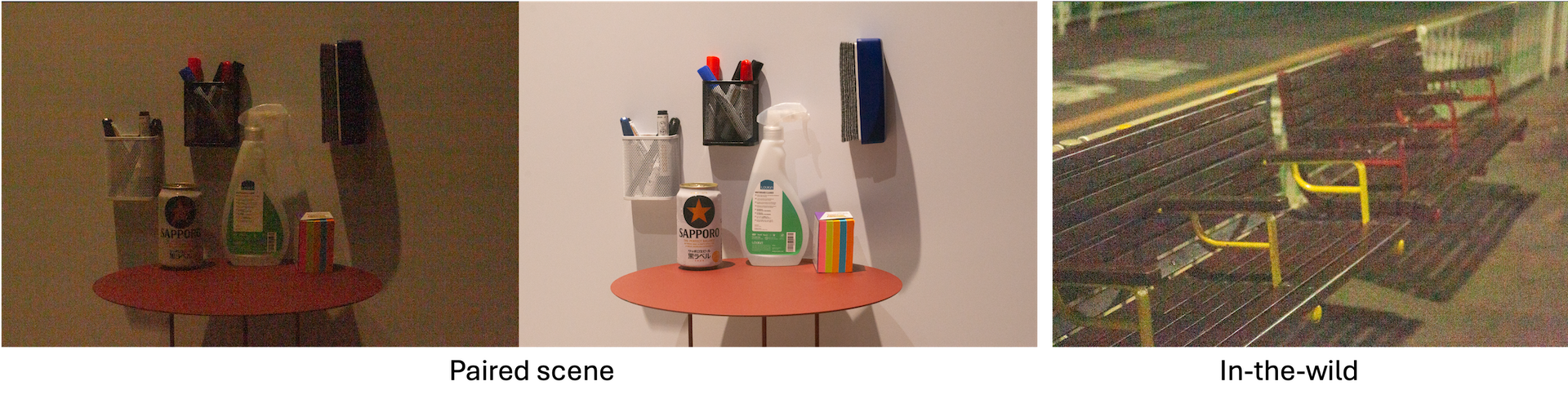}
    \caption{Data samples from the AIM 2025 RAW Image Denoising Challenge dataset.}
    \label{fig:samples}
\end{figure*}
\section{The AIM 2025 Real-World RAW Image Denoising Challenge}
\label{sec:challenge_detail}
The challenge encourages solution methods that perform precise noise synthesis and facilitate the training of denoising neural networks in a camera-agnostic manner. A total of $86$ teams participated in the challenge, of which $97$ teams submitted valid results in the final testing phase. 

\paragraph{Challenge Approaches} Participants are encouraged to approach this challenge from two key perspectives as shown in Figure~\ref{fig:rawdenoise}.
\begin{itemize}
    \item \textbf{Better noise modeling:} novel usage of noise profiles from multiple cameras to enhance the noisy image synthesis pipeline for self-supervised learning --- see Figure~\ref{fig:noise}.

    \item \textbf{Better denoising methodologies:} novel designs in network architectures, training strategies, or other techniques to achieve camera-agnostic RAW image denoising.
\end{itemize}


\begin{table*}[t]
    \centering 
    \caption{\textbf{AIM 2025 Real-World RAW Image Denoising Benchmark.} The best and second best results are in \textbf{bold} and \underline{underlined}, respectively (In the overall ranking, although MR-CAS and IPIU-LAB achieve the same average ranking score, MR-CAS outperforms IPIU-LAB in 3 out of the 5 metrics and is therefore ranked first).}
    \label{tab:results}
    \begin{tabular}{
        r  
        c
        c
        c
        c
        c
        c  
        c  
        c  
    }
        \toprule 
        
        \multirow{2}{*}{\textbf{Method}} &
        \multirow{2}{*}{\textbf{PSNR}$\uparrow$} &
        \multirow{2}{*}{\textbf{SSIM}$\uparrow$} &
        \multirow{2}{*}{\textbf{LPIPS}$\downarrow$} &
        \multirow{2}{*}{\textbf{ARNIQA}$\uparrow$} &
        \multirow{2}{*}{\textbf{TOPIQ}$\uparrow$} &
        \multicolumn{3}{c}{\textbf{Rank}} \\
        \cmidrule(lr){7-9} 
        
        & & & & & & \textbf{Overall} & \textbf{Fidelity} & \textbf{Perceptual} \\
        
        \midrule 
        MR-CAS~(\ref{sec:mrcas})     & \textbf{41.90} & \textbf{0.9633} & 0.2314 & 0.4615 & 0.2584 & 1 & 1 & 3 \\
        
        IPIU-LAB~(\ref{sec:ipiu}) &41.59	&0.9621	&0.2426	&\textbf{0.4698}	&\underline{0.2619} & 2 & 2 & 1 \\
        
        
        VMCL-ISP~(\ref{sec:vmcl})   & 41.15 & 0.9585 & 0.2443 & 0.4631 & \textbf{0.2671} & 3 & 6 & 2 \\
        
        HIT-IIL~(\ref{sec:hit})        & 41.52 & 0.9605 & \underline{0.2295} & 0.4374 & 0.2540 & 4 & 3 & 6 \\
        
        DIPLab~(\ref{sec:diplab})       & 41.23 & 0.9592 & \textbf{0.2182} & 0.4227 & 0.2567 & 5 & 4 & 4 \\
        
        MSA-Net~(\ref{sec:lucky})        & 41.13 & 0.9596 & 0.2523 & {0.4680} & 0.2576 & 6 & 5 & 5 \\
        
        
        MS-Unet~(\ref{sec:chaos})   & 40.82 & 0.9581 & 0.2506 & \underline{0.4684} & 0.2463 & 7 & 7 & 7 \\
        
        \bottomrule 
    \end{tabular}
\end{table*}

\paragraph{Dataset}
Participants are free to use any clean images of their choice for training data synthesis. For validation and benchmarking, We provide a dedicated dataset captured using four different DSLR cameras: Sony A7R IV, Sony A6700, Sony ZV-E10M2, and Canon 70D. These cameras feature CMOS sensors ranging from high-resolution full-frame to APS-C sizes. For each camera, the dataset comprises two types of scenes:
\begin{itemize}
    \item \textbf{Paired scenes}: Following the acquisition pipeline described in~\cite{wei2021physics}, for each scenario, noisy images are captured under three ISO levels ($800$, $1600$, and $3200$) and two digital gains ($100$ and $200$). Each noisy frame is paired with a clean ground-truth image obtained via a long-exposure shot at the sensor’s base ISO, while all other capture parameters were held constant to guarantee precise pixel-wise alignment. For each camera, the aforementioned captures are conducted across 10 distinct indoor scenes, which are evenly split into $50\%$ for validation and $50\%$ for testing.
    \item \textbf{In-the-wild scenes}: For each camera, noisy images are collected across $40$ scenarios, with $10$ used for validation and the remaining $30$ for testing. For each scenario, ISO levels are randomly selected from five settings ($800$, $1250$, $1600$, $3200$, and $6400$), with digital gains picking from the range $\left(10, 100\right)$. Most of the captures were conducted in outdoor environments, reflecting real-world conditions and providing diverse, challenging scenarios for accurate noise modeling and effective denoising. 
\end{itemize}

To support the formulation of precise noise synthesis pipelines, we also provide calibrated system gains and $50$ dark frames (\ie, captured w/o incident light) for each camera at each ISO level. Overall, this benchmark dataset serves as a robust foundation for participants to develop and evaluate their camera-agnostic RAW denoising solutions. We show sample images from our dataset in Figure~\ref{fig:samples}.

\paragraph{Evaluation protocol}
Both full-reference and no-reference image quality assessment metrics are employed to comprehensively evaluate the fidelity and perceptual restoration capabilities of each candidate method. Details are provided below:
\begin{itemize}
    \item \textbf{Metric:} PSNR, SSIM, and LPIPS~\cite{zhang2018perceptual} are employed as full-reference metrics on paired scenes, while ARNIQA~\cite{agnolucci2024arniqa} and TOPIQ~\cite{chen2024topiq} are used as no-reference metrics for in-the-wild scenes. Among these, PSNR and SSIM are computed directly on the predicted Bayer RAW images, and LPIPS, ARNIQA, and TOPIQ\footnote{Implementations of LPIPS, ARNIQA, and TOPIQ are sourced from PyIQA: \url{https://github.com/chaofengc/IQA-PyTorch}} are applied after a basic image signal processing (ISP) pipeline. Images are center-cropped to $512\times512\times4$ (\ie, corresponding to $1024\times1024\times3$ ISP-processed sRGB patches) in the development phase, and $1024\times1024\times4$ in the final testing phase.
    \item \textbf{Final ranking method:} Participants are first ranked independently for each metric, and the relative rankings are recorded as ranking scores. Subsequently, average ranking scores are computed in three categories: overall, fidelity (\ie, PSNR and SSIM), and perceptual (LPIPS, TOPIQ, and ARNIQA). A lower average ranking score indicates better performance.
\end{itemize}

\paragraph{Efficiency} We propose the following efficiency requirements to constraint the model solutions and study realistic denoising applications:
\begin{itemize}
    \item Maximum 15 million parameters for the neural network.
    \item MACs for the input shape of (1, 4, 512, 512) shall be less than 150 GMacs.
    \item Ensembles of multiple models are not allowed.
\end{itemize}

\section{Challenge Results}
\label{subsec:results}

The final results of the competition are listed in \Cref{tab:results}. The winner, MR-CAS~(\ref{sec:mrcas}), proposes a random
masking strategy consistent with Masked Autoencoder to improve the generalization capabilities of the models. Most of the proposed solutions use NAFNet~\cite{chen2022simple} as the baseline, and propose incremental improvements such as novel attention mechanisms. However, we can see the biggest benefits in data synthesis and training strategies. We provide a summary of the implementation details in Table~\ref{tab:summary}, and results using the public validation set (Codabench site) in Table~\ref{tab:validation}.

\begin{table}
    \centering
    \caption{Implementation details summary.}
    \resizebox{\linewidth}{!}{
    \begin{tabular}{l c c c c c c}
        \toprule
        Method & Input & Time (h) & E2E & Extra Data & Params. (M) & GPU  \\
        \midrule
        FrENet & 512 & 24 & Yes & No & 5 & 3090 \\
        HIT-IIL & 512 & 120 & Yes & Yes & 13.93 & A6000 \\
        MSA-Net & 512 & 41  & Yes & No & 4.89 & 3090 \\
        MS-Unet & 512 & 28 & Yes & No & 8.13 & 2 x 4090 \\
        DIPLab & 512 & 20 & Yes & No & 14.02 & A100\\
        VMCL-ISP & 256 & 35 & Yes & No & 13.7 & 8 x 4090 \\
        \bottomrule
    \end{tabular}
    }
    \label{tab:summary}
\end{table}

\begin{table}[t]
    \centering
    \caption{Summary results using the challenge validation set.}
    \resizebox{0.9\linewidth}{!}{
    \begin{tabular}{l c c c c}
         \toprule
         Method & PSNR & SSIM & Params. (M) & GMACs  \\
         \hline
         Input  & 20.298 & 0.1553 & - & - \\
         FrENet & 42.906 & 0.9683 & 14.92 & 93.93 \\
         DIPLab & 42.327 & 0.9647 & 14.02M & 142.94 \\
         MS-Unet & 42.011 & 0.9639 & 8.13 & 67.36 \\
         \bottomrule
    \end{tabular}
    }
    \label{tab:validation}
\end{table}

\section{Challenge Methods}
\label{subsec:methods}

\begin{disclaimerbox}
In the following Sections, we describe the top challenge solutions -- each was checked manually by the organizers to ensure fairness.

Note that the method descriptions were provided by each team as their contribution to this report.
\end{disclaimerbox}

\subsection{Image denoising with random mask}
\label{sec:mrcas}


\begin{center}

\vspace{2mm}
\noindent\emph{\textbf{MR-CAS}}
\vspace{2mm}

\noindent\emph{Gaozheng Pei$^{1}$, Ke Ma$^{1}$, Chengzhi Sun$^{1}$, Qianqian Xu$^{2}$, Qingming Huang$^{1}$}

\vspace{2mm}

\noindent\emph{$^{1}$University of Chinese Academy of Sciences\\$^{2}$Institute of Computing Technology, Chinese Academy of Sciences}

\vspace{2mm}

\noindent{\emph{Contact: \url{peigaozheng23@mails.ucas.ac.cn}}}

\end{center}

We tested three model architectures, including U-net, Restormer, and NAFNet. We modified these three model structures to precisely meet the parameter and computational requirements of the Challenge. Experimental results showed that U-net performed the worst, Restormer was in the middle, and NAFNet achieved the best performance. Therefore, we chose to adopt NAFNet. 
                                         
To address the weak generalization capability of self-supervised denoising methods, we employed a random masking strategy consistent with Masked Autoencoder. This approach enables the model to genuinely understand image content, thereby allowing its denoising capability to generalize to unseen noise types.

\paragraph{Global Method Description}
The existing deep learning denoising methods have a critical issue—poor generalization capability, which is particularly severe in self-supervised raw image denoising because real-world noise modeling varies across different camera types. To enhance the model's generalization capability, we need the model to truly understand the image content.

Inspired by \cite{he2022masked,chen2023masked} while aiming to enhance versatility without modifying the network architecture, unlike \cite{chen2023masked}, we adopt the same strategy as \cite{he2022masked} by performing random masking at the image level. For the model architecture, we employ NAFNet \cite{chen2022simple} and adjust the number of intermediate blocks along with the feature dimensions to ensure the model's parameters and computational complexity meet the challenge requirements. 

To further enhance the model's generalization capability, we incorporated additional datasets for training. We observed a discrepancy between the resolution during final testing and training. To mitigate the impact of resolution variation on denoising performance, we implemented progressive learning by training the network with gradually increasing image sizes from 128x128 to 256×256 and finally 1024×1024. For data augmentation, we applied random rotations to the images at four specific angles. Our approach follows a multi-stage training paradigm, where each subsequent stage initializes with the best-performing weights from the previous training stage.

\paragraph{Implementation details}
\begin{itemize}
    \item \textbf{Architecture:} We use NAFNet with a feature dimension (width) of 32, middle\_blk\_num set to 5, enc\_blk\_nums as [2, 2, 4, 4], and dec\_blk\_nums as [2, 2, 2, 2].
    \item \textbf{Optimizer and Learning Rate:}  We employ the AdamW optimizer with an initial learning rate of 3e-5 and utilize CosineAnnealingLR for learning rate scheduling.
    \item \textbf{GPU:} The GPU we used is NVIDIA GeForce RTX 4090 24GB Memory.
    \item \textbf{Datasets:} In addition to the SID dataset, we incorporated supplementary datasets including ELD \cite{wei2021physics}, low-light
 raw image dataset captured with a Nikon camera \cite{prabhakar2023few}.
    \item \textbf{Training Time:} The model was trained for approximately three weeks using 4-8 NVIDIA RTX 4090 GPUs.
    \item \textbf{Training Strategies:} We implemented multi-stage training with 500 epochs per stage, which can be divided into four main phases. Each subsequent stage initializes with the final weights from the previous training phase. The first stage uses L1 loss with the SID dataset. In the second stage, we incorporate additional datasets while maintaining L1 loss. The third stage introduces random masking of input data while continuing to use L1 loss. The final stage employs a combined training approach using both Charbonnier loss and L1 loss.
    \item \textbf{Data Augmentation:} We perform random image rotations with equal probability among three angular options (90°, 180°, 270°, 360°). In the final two training stages, we employed random masking with a hybrid ratio of 75\% and 50\%, using a patch size of 16×16 pixels. Each batch randomly masks half of the samples.
    \item \textbf{Loss Function:} We exclusively employed a hybrid loss function in the final stage, combining L1 loss (weight: 1.0) and Charbonnier loss (weight: 0.1).
\end{itemize}

\newpage

\subsection{Efficient RAW Image Denoising with Adaptive  Frequency Modulation}
\label{sec:ipiu}


\begin{center}

\vspace{2mm}
\noindent\emph{\textbf{IPIU-LAB}}
\vspace{2mm}

\noindent\emph{Yiqing Wang, Jing He, Kexin Zhang, Licheng Jiao, Lingling Li, Wenping Ma}

\vspace{2mm}

\noindent\emph{Intelligent Perception and Image Understanding Lab, Xidian University\\
Intelligent Perception and Image Understanding Lab, Xidian University}

\vspace{2mm}

\noindent{\emph{Contact: \url{24171213882@stu.xidian.edu.cn}}}

\end{center}

We used FrENet \cite{jiao2025efficientrawimagedeblurring} in the challenge, adjusted its parameters, fine-tuned it under the competition's model constraints, and achieved a validation set PSNR of 48.818. We did not test existing methods, focusing instead on developing and optimizing our own models for RAW image denoising without comparative experiments.

The Frequency Enhanced Network (FrENet) is a frequency-domain framework for raw-to-raw deblurring. It integrates spatial and frequency processing through a U-Net architecture, featuring an Adaptive Frequency Positional Modulation (AFPM) module for dynamic frequency adjustment and frequency skip connections to preserve high-frequency details. We adapted it to RAW denoising by fine-tuning the modulation range of AFPM and optimizing the network depth to meet size constraints, achieving efficient denoising performance.

\subsubsection{Global Method Description}

\paragraph{Efficient RAW Image Denoising with Adaptive  Frequency Modulation}

\begin{figure}[t]
    \centering 
    \includegraphics[width=3in]{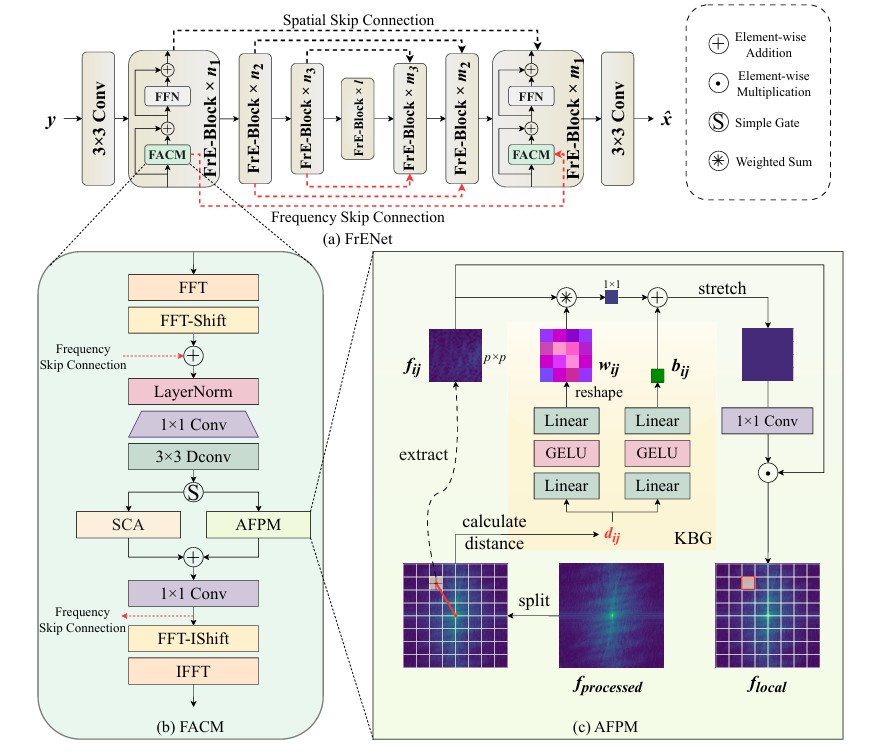}
    \caption{The model framework diagram of FrENet.} 
    \label{fig:FrENet} 
\end{figure}

FrENet employs a U-shaped structure with encoder, bottleneck, and decoder. Its core is enhancing feature expression via frequency domain analysis while maintaining efficiency. The 4-channel RAW input (Bayer pattern) is mapped to high-dimensional features through an initial 3×3 convolution. The encoder has L levels with multiple FrE-Blocks (each combining FACM and FFN), where feature resolution halves and channels double with each level to extract frequency features. Symmetric to the encoder, the decoder’s L levels restore resolution via upsampling, supplement details using encoder spatial/frequency skip connections, and halve channels gradually. Finally, a 3×3 convolution maps decoder outputs back to 4 channels, generating the denoised RAW image. The model framework diagram of FrENet is shown in fig\ref{fig:FrENet}.

\paragraph{FACM: Frequency Adaptive Context Module}
As the core frequency-domain processing sub-module in FrE-Block, FACM operates progressively: Input spatial features are transformed to the frequency domain via FFT with FFT-Shift centering zero frequency (decoder blocks further fuse encoder skip connections for initial \(f_{freq}\)). Real/imaginary parts of \(f_{freq}\) are channel-concatenated and LayerNorm-normalized. After 1×1 convolution (channel fusion), 3×3 depth-wise convolution (local frequency correlations), and SimpleGate activation, intermediate \(f_{processed}\) is generated. A dual-branch enhancement follows: local branch (AFPM) splits \(f_{processed}\) into patches, generating position-sensitive modulation kernels/biases via KBG based on patch-center distance for adaptive frequency adjustment; global branch (SCA) uses adaptive average pooling and 1×1 convolution for channel attention calibration. Fused local-global features are integrated via 1×1 convolution, then converted back to spatial domain via FFT-IShift and IFFT.

\paragraph{FFN: Feed-Forward Network}
The FFN, focusing on non-linear spatial enhancement of spatial features, adopts Restormer's efficient structure: 1×1 convolution expands channels, 3×3 depth-wise convolution captures spatial correlations, and content-aware gating (via element-wise multiplication with GELU-activated features from another branch) is used, with channel compression and residual connections. It complements FACM's output: FACM provides frequency-optimized base features, while FFN strengthens spatial detail expression via non-linear transformation. Together, they form a "frequency-spatial" bidirectional optimization loop, retaining frequency-domain sensitivity to noise/details and enhancing spatial-domain local texture modeling.

\subsubsection{Dataset and Preprocessing}
We used the Sony subset of the SID dataset as the training set. To accurately simulate noise characteristics in real shooting scenarios and enhance the model's generalization ability across different devices and shooting parameters, the data preprocessing involves three core stages\cite{li2025noise}: clean image construction, noise sample generation, and noisy image synthesis. The specific steps are as follows:
\begin{itemize}
\item Random Selection of Shooting Parameters and Device Information. ISO and camera model are randomly selected from presets. ISO affects noise, the model determines sensor noise traits and effective area. Subsequent processing stays within this area to avoid edge invalid pixels and ensure data validity.

\item Preprocessing of Clean Images. Original clean image RAW data is read, with the sensor's white level and black level extracted. Single-channel Bayer array data is converted to 4-channel RGGB format, with normalization and outlier clipping. The image is randomly cropped into multiple fixed-size sub-blocks to enhance training data diversity. Finally, sub-blocks are converted to a model-suitable tensor format, with preset data augmentation applied to improve generalization.

\item Preprocessing of Noise Frames. Dark frames matching the camera model and ISO are randomly selected, with their white and black levels extracted. Dark frames are corrected by removing spatially uneven dark current interference (dark shading) and subtracting the black level, yielding "signal-independent noise" (only sensor inherent noise). Corrected dark frames are cropped to the effective imaging area and converted to 4-channel RGGB format, to provide benchmark noise samples for subsequent synthesis.
\end{itemize}

Through the above pipeline, the preprocessed dataset can generate "noisy image-clean image" sample pairs with real noise characteristics, laying a data foundation for the model to learn noise suppression strategies in different scenarios.

\begin{itemize}

\item Discuss \textbf{Efficiency} of your method (MACs, FLOPs, runtime in ms)

\end{itemize}

\subsubsection{Implementation details}


\begin{itemize}
    \item \textbf{Framework:} PyTorch.
    \item \textbf{Optimizer and Learning Rate:} Optimizer is Adam, Learning Rate is 0.001 and learning rate decay strategy is cosine.
    \item \textbf{GPU:} Training: 1 $\times$ NVIDIA GeForce RTX 3090 24G. Inference: 1 $\times$ Tesla V100-SXM2-32GB.
    \item \textbf{Datasets:} The Sony subset of the SID dataset.
    \item \textbf{Training Time:} Training for 2000 epochs takes approximately 48 hours.
    \item \textbf{Training Strategies:} Fine-tuning.
\end{itemize}
\newpage
\subsection{PMNNP: A hybrid noise modeling}
\label{sec:vmcl}


\begin{center}

\vspace{2mm}
\noindent\emph{\textbf{VMCL-ISP}}
\vspace{2mm}

\noindent\emph{Hansen Feng$^1$, Zhanyi Tie$^1$, Ziming Xia$^1$, Lizhi Wang$^2$}

\vspace{2mm}

\noindent\emph{Beijing Institute of Technology\\Beijing Normal University}

\vspace{2mm}

\noindent{\emph{Contact: \url{hansen97@outlook.com}}}

\end{center}

Our method focuses on accurate noise modeling, which is critical for extreme low-light raw image denoising. We propose PMNNP, a hybrid noise modeling strategy that extends SFRN+DSC~\cite{zhang2021rethinking} from PMN~\cite{PMN} and incorporates the PNNP* formulation~\cite{feng2023physics}. The noise model is calibrated using the official dark frame dataset. For shot noise, frame-wise noise and band-wise noise, we adopt the modeling approach of PNNP. For pixel-wise noise, we blend synthetic noise generated by PNNP with real pixel-wise noise extracted from dark frames.

\begin{figure}[t]
	\centering
	\includegraphics[width=\linewidth]{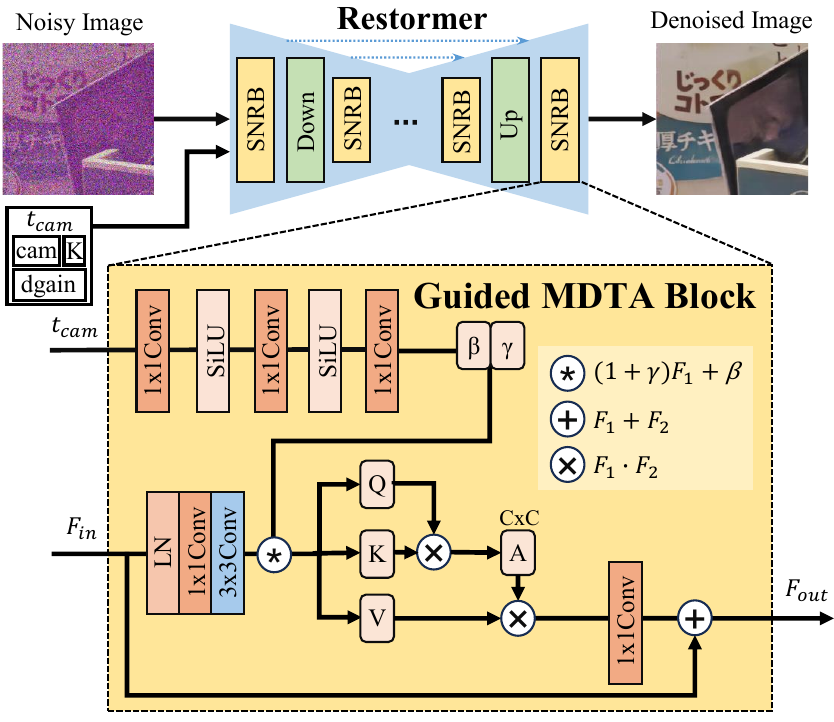}
	\caption{Team VMCL-ISP. Overview of the proposed PMNNP (Restormer).}
	\label{fig:network-vmcl}
\end{figure}

Our network builds upon Restormer~\cite{Restormer} as shown in Figure~\ref{fig:network-vmcl}. We introduce two modifications: (1) reducing the original block count for improved efficiency, and (2) adding a guidance branch inspired by YOND~\cite{feng2025yond}. The guidance branch adjusts network behavior according to the camera type, analog gain and digital gain, enabling robust noise adaptation across different sensors.

\begin{figure}[t]
	\centering
	\includegraphics[width=\linewidth]{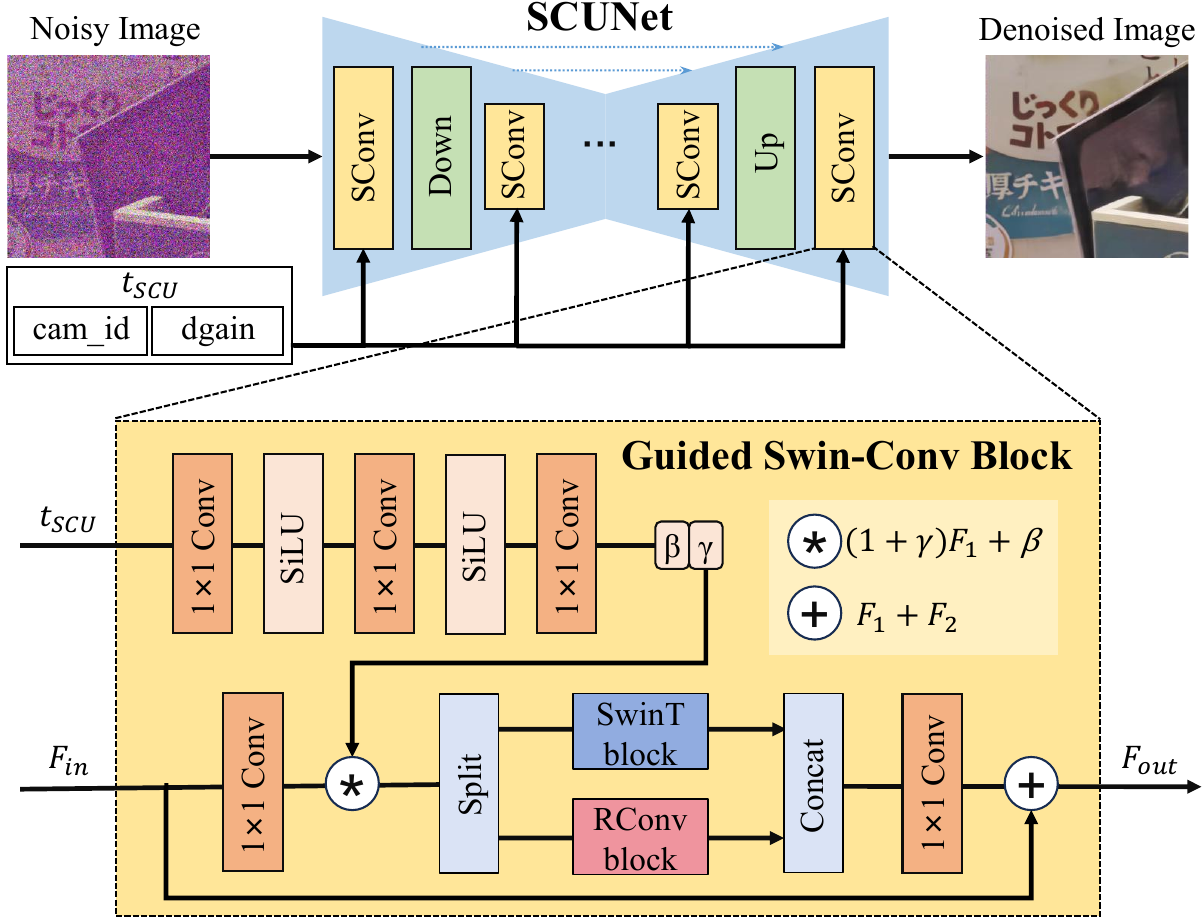}
	\caption{Team VMCL-ISP. Overview of the proposed PMNNP (SCUNet).}
	\label{fig:network-vmcl2}
\end{figure}

\paragraph{Comparisons on the Validation Set}
We evaluate a range of representative noise modeling approaches on the validation set, with results summarized in Table~\ref{tab:results-vmcl}. All methods share the same backbone and training schedule, thus the primary differences lie in the noise modeling and the use of noise parameters.

\begin{table}
	\centering
	\begin{tabular}{l c c c c}
		\toprule
		Method & PSNR & SSIM & Params (M)  \\
		\midrule
		Baseline images & 20.30 & 0.1553 & -\\
		\midrule
		VST+AWGN & 41.11 & 0.9417 & \textbf{13.947}\\
		VST+PNNP & 41.50 & 0.9597 & \textbf{13.947}\\
		VST+PMNNP & 41.44 & 0.9599 & \textbf{13.947}\\
		\midrule
		kSigma+SFRN+DSC & 42.11 & 0.9610 & 13.962\\
		kSigma+PNNP & 42.24 & 0.9613 & 13.962\\
		kSigma+PMNNP & 42.19 & 0.9613 & 13.962\\
		\midrule
		SFRN+DSC & 42.64 & 0.9641 & 13.962\\
		PNNP & 42.59 & 0.9635 & 13.962\\
		PMNNP & 42.69 & 0.9640 & 13.962\\
		PMNNP finetune & \textbf{42.75} & \textbf{0.9647} & 13.962\\
		\bottomrule
	\end{tabular}
	\caption{Summary results on validation set (codabench) of VMCL. All methods employ the same SCUNet backbone, thus their computational complexity is similar.}
	\label{tab:results-vmcl}
\end{table}

Based on how noise parameters are utilized, these methods can be classified into three categories:
\textbf{VST-based} methods aim to transform arbitrary camera noise into additive white Gaussian noise via variance-stabilizing transforms~\cite{feng2025yond}; 
\textbf{kSigma-based} methods normalize Poisson-Gaussian noise to simplified data mapping~\cite{Yuzhi}; 
\textbf{Non-transform} methods denoise directly on noisy raw images without any transformation. 

According to our observation, the instability of physical imaging environments often leads to misalignment between calibrated noise parameters and real-world conditions. As a result, transform-based methods may break their underlying assumptions in practice. These results underscore the persistent challenge of properly incorporating noise parameters under extreme low-light condition.

From the perspective of noise modeling, PNNP achieves the best performance among transform-based methods, while PMNNP performs best among non-transform methods. A detailed comparison of denoising results reveals notable preferences across different noise modeling methods. PNNP tends to preserve fine details but may leave residual noise or artifacts. In contrast, SFRN+DSC produces clean results but often oversmooths low-SNR textures. PMNNP strikes a balance between the robustness of SFRN+DSC and the detail preservation of PNNP, thereby delivering superior overall performance.

\paragraph{Implementation details}

We implement our method using PyTorch and train all models on 8 NVIDIA RTX 4090 GPUs.
We adopt Restormer~\cite{Restormer} as the backbone and reduce the number of MDTA blocks in each layer to [1, 2, 4, 8]. The channel dimensions of guidance branch are aligned with the corresponding backbone blocks.
The training set of the SID dataset~\cite{Chen_2018_CVPR} is used exclusively as ground truth. Due to noticeable residual noise in high-ISO images, we apply a blind raw denoising method~\cite{feng2025yond} to clean these images before training.
Training is performed in 3 stages, each for 200 epochs, with a total training time of approximately 35 hours. In the first stage, we adopt the PNNP noise model to enable robust learning across arbitrary ISO levels. In the second stage, we fine-tune the model using the proposed PMNNP to align with real noise characteristics. In the final stage, we introduce an additional SSIM loss to enhance detail preservation, while L1 loss is used throughout all stages. The optimizer is AdamW with a cosine annealing learning rate schedule. The initial learning rates are set to 2e-4, 1e-4, and 1e-4, respectively.
During inference, we divide large images into overlapping 256*256 patches, perform denoising on each patch, and then blend them back into the full image. Notably, to prevent highlight color shifts, we modify the default clip upper bound in the official code from 1 to 2.

\begin{figure}[t]
	\footnotesize
	\setlength\tabcolsep{3pt}
	\centering
	\begin{tabular}{ccc}
		Canon70D Scene-50 & Canon70D ISO-3200 & SonyZVE10M2 ISO-800\\
		{\includegraphics[width=0.31\linewidth]{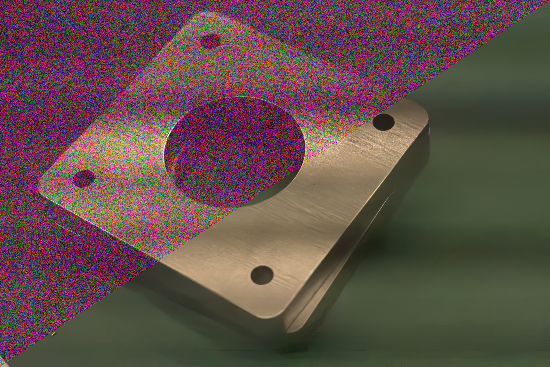}} &
		{\includegraphics[width=0.31\linewidth]{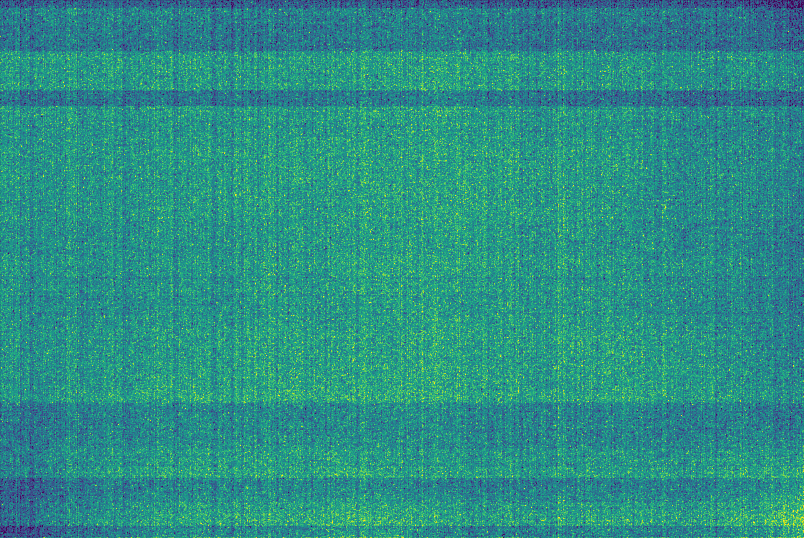}} &
		{\includegraphics[width=0.31\linewidth]{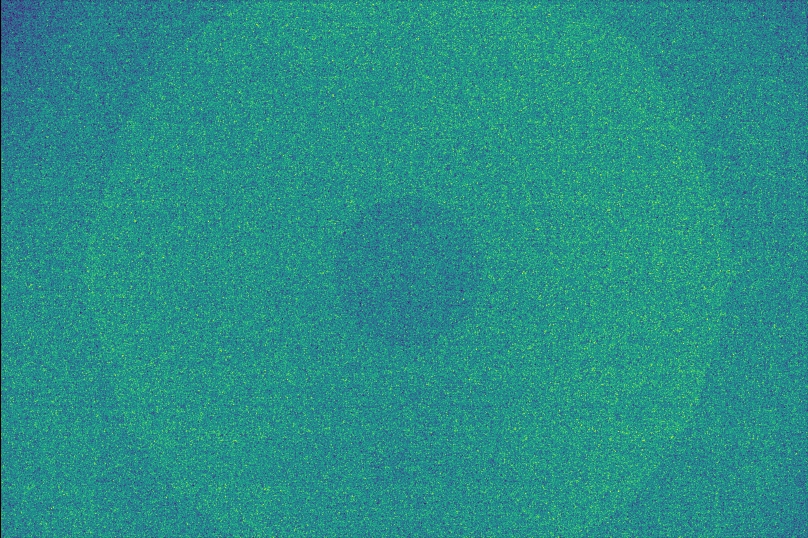}} \\
		(a) Flicker Banding & (b) Overheating FPN & (c) Lens Compensation
	\end{tabular}
	\caption{Examples of data defects observed in the official dataset}
	\label{fig:defects}
	\vspace{-6pt}
\end{figure}

\paragraph{Discussion on Data Quality}
As shown in Figure~\ref{fig:defects}, we identify a few data defects in the official dataset that may compromise the consistency of noise modeling and evaluation. In the dark frames, pattern noise introduced by sensor overheating and compensation signals triggered by lens mechanisms are observed. These artifacts, however, do not appear in the actual test scenes. In the short-exposure inputs, flicker banding occurs, likely due to a mismatch between indoor lighting frequency and exposure time. Such banding is not expected in the long-exposure ground truth.

We suggest considering the data acquisition protocols proposed in PMN~\cite{PMN} as a potential way to mitigate some of these issues in future releases.

\newpage
\subsection{Scaling Up Data for Better Denoising}
\label{sec:hit}


\begin{center}

\vspace{2mm}
\noindent\emph{\textbf{HIT-IIL}}
\vspace{2mm}

\noindent\emph{Mingyang Chen, Renlong Wu, Junyi Li, Zhilu Zhang, Wangmeng Zuo}

\vspace{2mm}

\noindent\emph{Faculty of Computing, Harbin Institute of Technology}

\vspace{2mm}

\noindent{\emph{Contact: \url{youngmchan269@gmail.com}}}

\end{center}

We enhance real-world RAW denoising by leveraging more higher-quality training data.
Specifically, beyond employing clean images from SID dataset\cite{Chen_2018_CVPR}, we collect 1200 ones in indoor and outdoor scenes with a Sony camera, where low-quality samples are filtered based on no-reference image quality assessment metrics.
We employ NAFNet\cite{chen2022simple} with 13.93M parameters as the denoising network.
The inference cost on $4\times512\times512$ images is 138.59 GMacs.

\paragraph{Global Method Description}
We adopt NAFNet~\cite{chen2022simple} as the denoising network.
We train the model on synthetic data, with noise synthesized according to the provided noise model parameters.
We find that the size and quality of training data have a crucial impact on performance.
Thus, beyond employing 231 clean long-exposure images from SID dataset\cite{Chen_2018_CVPR}, we collect 1,200 long-exposure images in indoor and outdoor scenes with a Sony camera.
%
To ensure high data quality, we automatically filter 20\% low-quality images baed on the averaged score of no-reference IQA metrics (\ie, Laplacian Variance~\cite{pertuz2013analysis}, BRISQUE~\cite{mittal2012no}, and NIQE~\cite{mittal2012making}).
%
%
%
We utilize $\ell_1$ loss as the loss function.
During training, we randomly crop patches and augment them with random flips. 
The patch size is progressively set from $256 \times 256$ to $1024 \times 1024$.

\paragraph{Implementation details}

\begin{itemize}
    \item \textbf{Framework:} PyTorch.
    \item \textbf{Optimizer and Learning Rate:} We use the Adam optimizer. The initial learning rate is set to $1 \times 10^{-4}$ and decayed to $1 \times 10^{-7}$.
    \item \textbf{GPU:} We conduct experiments on a NVIDIA RTX A6000 GPU. We use about 44GB of GPU memory.
    \item \textbf {Datasets:}  We use 231 long-exposure images and 1200 self-collected real-world ones as the clean images.
    
    \item \textbf{Training Strategies:} 
    We utilize $\ell_1$ loss as the loss function.
    We randomly crop patches and augment them with random flips. 
    The patch size is progressively set from $256 \times 256$ to $1024 \times 1024$.
    
    \item \textbf{Efficiency Optimization Strategies:} 
    We build upon NAFNet~\cite{chen2022simple}, where the number of base channel is set to 48.
    The encoder consists of 2, 4 and 6 NAFNet blocks for each scale, respectively.
    The decoder consists of 2, 2 and 2 NAFNet blocks for each scale, respectively.
\end{itemize}

\subsection{Bayer Group Convolution for Raw Image Processing}
\label{sec:diplab}


\begin{center}

\vspace{2mm}
\noindent\emph{\textbf{DIP Lab}}
\vspace{2mm}

\noindent\emph{Jaeseong Yu, Hongjae Lee, Myungjun Son, and Seung-Won Jung}

\vspace{2mm}

\noindent\emph{Korea University}

\vspace{2mm}

\noindent{\emph{Contact: \url{jsyu624@korea.ac.kr}}}

\end{center}

We design a lightweight Bayer Group Convolution (BGC) module that incorporates CFA structure into the kernel design. We show that BGC can be integrated into existing networks to improve both accuracy and efficiency.

\noindent\textbf{Our Contributions are as follows:}
\begin{enumerate}[label=\arabic*. , leftmargin=10pt, itemsep=2pt]
    \item \textbf{CFA-aware Bayer Group Convolution.} 
          We introduce BGC for Bayer and generic \(N\times N\) CFA patterns.

    \item \textbf{Plug-and-play compatibility.}
          BGC enhances PSNR without increasing MACs, when applied to the first and last layers of the baseline.
\end{enumerate}

\begin{figure}[!t]
  \centering
  \includegraphics[width=.96\linewidth]{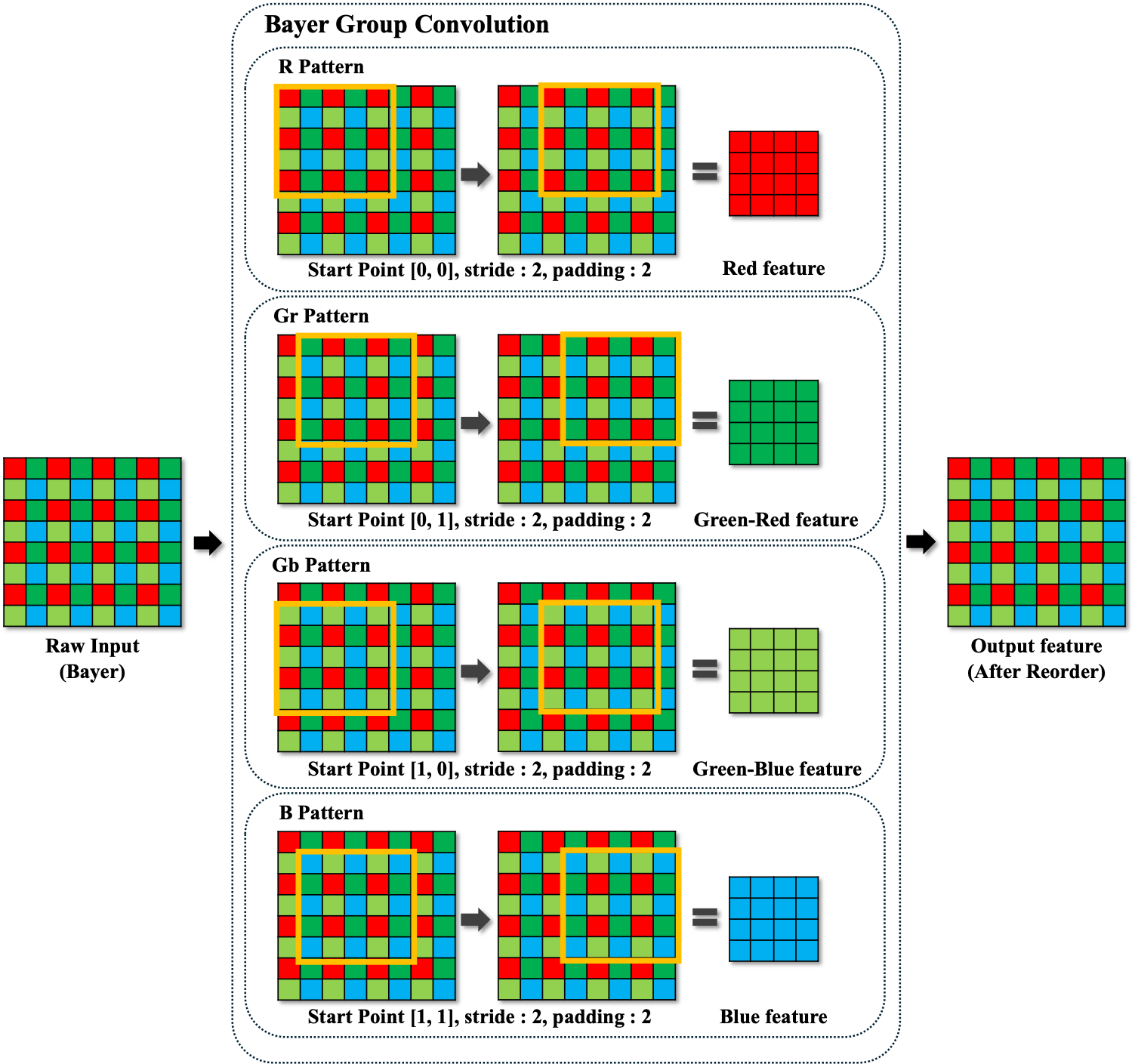}
  \caption{Schematic of the Bayer group convolution block. }
  \label{fig:bgc_operation}
\end{figure}

\begin{figure*}[t]
    \centering
    \includegraphics[width=.76\linewidth]{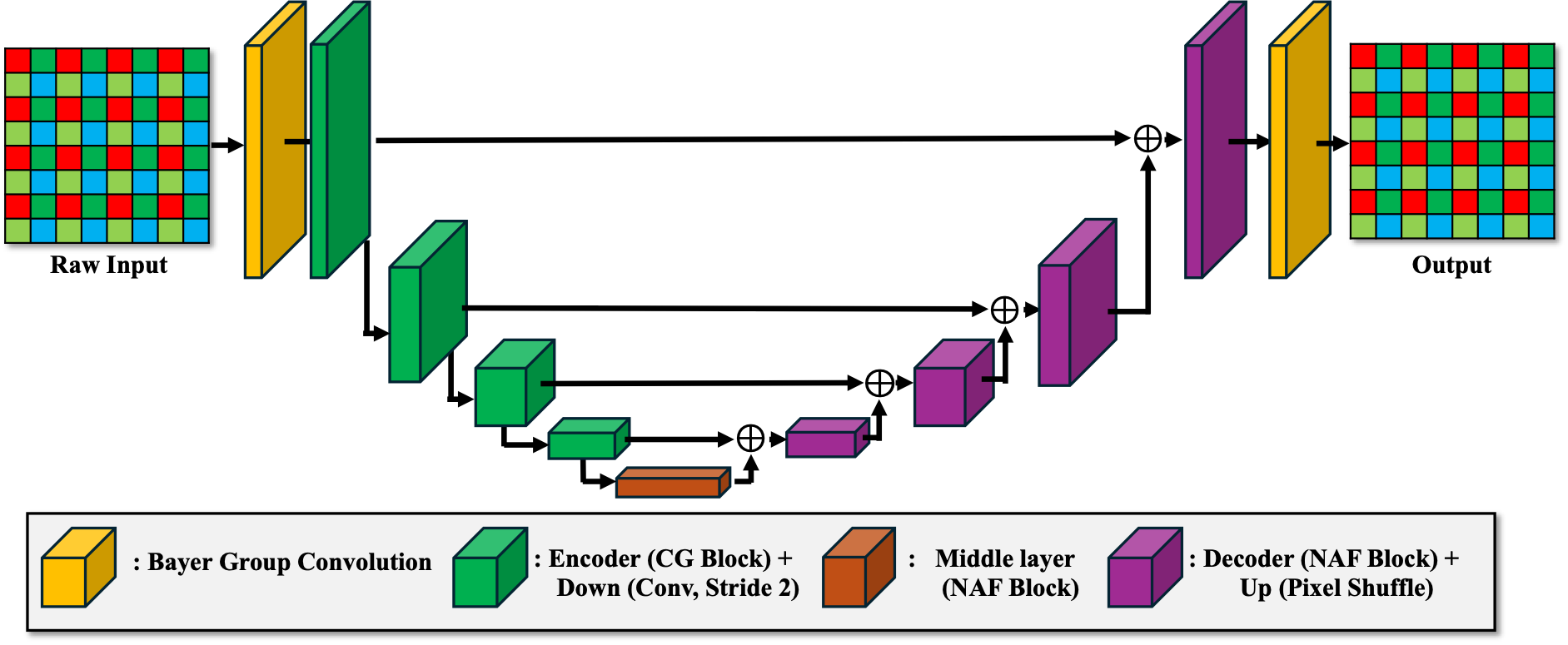}
    \caption{Overview of the proposed BGC blocks applied to the baseline~\cite{ghasemabadi2024cascadedgaze} by DIPLab.}
    \label{fig:diplab}
\end{figure*}

\subsubsection{Method Description}
We propose a BGC module that explicitly encodes the sensor’s CFA structure to effectively leverage the unique characteristics of raw images. BGC operates in the encoder stage and can be seamlessly integrated into existing pipelines as a lightweight plug-in, enhancing performance without added complexity.

\paragraph{Color filter array and convolution challenges}
In camera systems, image data are captured by an image sensor. Although the sensor type determines attributes such as resolution and sensitivity, the component most pertinent to this study is the CFA. During spatial sub-sampling, the CFA specifies which color filter is placed at each pixel location, and the resulting image response is influenced by the color array pattern. The most common patterns are the 2$\times$2 Bayer array and the 4$\times$4 Quad-Bayer array. A critical issue when performing convolution on CFA data is that the color information represented by each kernel weight depends on its relative position within the pattern.

\subsubsection{BGC : pattern aware feature extraction}

To exploit this pattern-dependent characteristic, we introduce BGC that partitions the spatial domain according to the CFA's N$\times$N period and performs independent convolutions for each group, enabling color-pattern-aware feature learning. In BGC, computation begins by decomposing the raw input into \(N^{2}\) sub-tensors,~\ref{fig:bgc_operation} each aligned with a specific position in the CFA period—an operation. These sub-tensors are convolved in parallel with dedicated kernels, whose weights are updated to reflect the color statistics of their respective CFA locations. The resulting feature maps are subsequently concatenated and reordered to recover the original CFA layout, preserving spatial coherence while enriching the representation with color-aware features. 

Because BGC makes the CFA pattern explicit, the network no longer needs to learn the Bayer topology implicitly; each kernel instead specializes in a single color channel. This pattern-aware design yields improved representational power on real raw data without increasing the parameter count and stabilizes the optimization of downstream modules.

\subsubsection{Implementation details}

Experiments and validation were conducted using the environment provided by the AIM Raw Denoise Challenge.

\paragraph{Datasets}
The training data were constructed exclusively from the Sony subset of the See-in-the-Dark (SID) dataset, as introduced by Chen~\etal~\cite{Chen_2018_CVPR}. Each long-exposure (raw-long) frame was regarded as a noise-free reference. Poisson (shot) noise was added and blended with dark-frame patterns that depend on ISO and digital gain to synthesize the corresponding short-exposure (raw-short) image Li~\etal~\cite{li2025noise}. All resulting pairs were randomly cropped to 256$\times$256 patches.

\paragraph{Baseline model}
We use CascadeGaze Net~\cite{ghasemabadi2024cascadedgaze} as the baseline model. To meet the constraints of the challenge, the encoder uses 1, 1, 2, and 4 CascadedGaze blocks across its four stages. The bottleneck consists of 4 NAF blocks, and each of the four decoder stages has 2 NAF blocks. We set the width of the network to 28. Both the initial and final layers use a 5$\times$5 BGC.

\paragraph{Final network description}
We implemented CascadedGaze Net with BGC using PyTorch. Training was carried out on two NVIDIA A100 GPUs. Following the standard protocol in Ghasemabadi,~\etal~\cite{ghasemabadi2024cascadedgaze}, we adopted the AdamW optimizer with $\beta = (0.9, 0.9)$ and zero weight decay. The initial learning rate was set to 1e-3 and subsequently reduced by a cosine-annealing schedule over 500 epochs, with a minimum learning rate of 1e-7. The network was optimized using PSNR loss.
\begin{figure*}[t]
    \centering
    \includegraphics[width=0.8\textwidth]{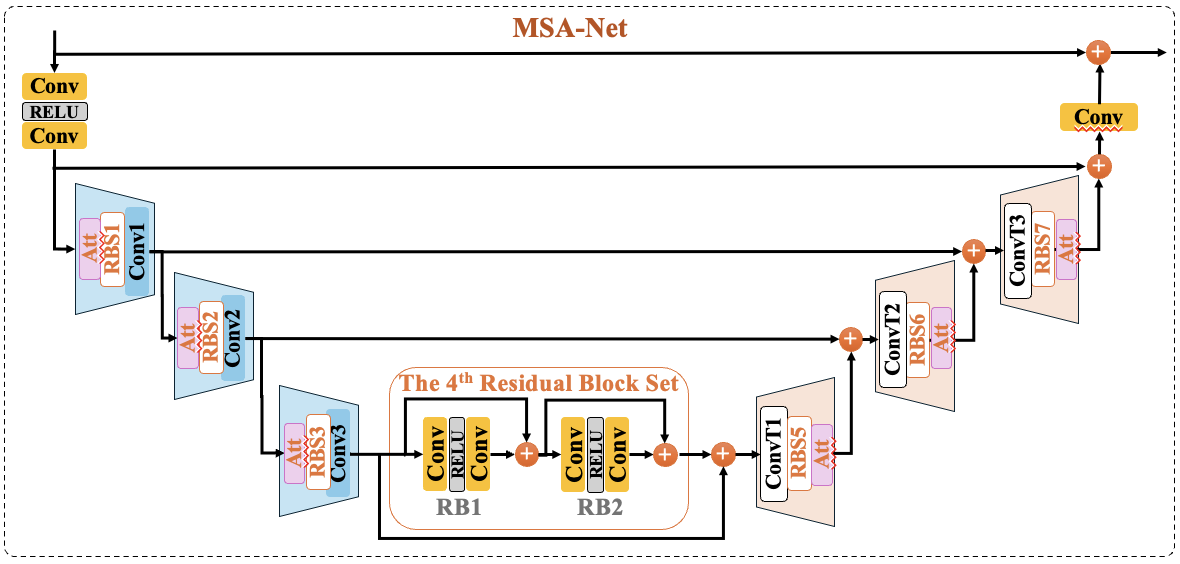}
    \caption{Illustration of the proposed solution MSA-Net.}
    \label{fig:MSA-Net}
\end{figure*}

\subsection{Multi-Scale Attention guided raw image de-
noising network (MSA-Net)}
\label{sec:lucky}


\begin{center}


\noindent\emph{Jingyi Xu}

\vspace{2mm}

\noindent\emph{Beihang University, Beijing, China}

\vspace{2mm}

\noindent{\emph{Contact: \url{jingyixu@buaa.edu.cn}}}

\end{center}

\begin{table}
    \centering
    \resizebox{\linewidth}{!}{
    \begin{tabular}{l c c c c}
         \toprule
         Method & PSNR & SSIM & Params & GMACs  \\
         \hline
         Baseline images & 41.607 & 0.9593 & - & - \\
         MSA-Net     & 42.182 & 0.9646 & 4.89MB & 109.26GMac \\
         MSA-Net-D   & 42.481 & 0.9674 & 4.89MB & 109.26GMac \\
         \bottomrule
    \end{tabular}
    }
    \caption{Summary results of MSA-Net using the validation set.}
    \label{tab:results-MSA-Net}
\end{table}

Our network design is based on a key observation regarding the clean and noisy raw image pairs provided in this track: the degradation levels of the 1st and 4th channels are significantly lower than those of the 2nd and 3rd channels, with an average PSNR difference of about 1 dB. This indicates that processing all 4-channel raw images simultaneously using a baseline network may restrict the denoising performance for the 2nd and 3rd channels, as the relatively clean information from the 1st and 4th channels cannot be effectively utilized to assist in restoring the more degraded ones.

To address this issue, we propose to add a Multi-Scale Attention (MSA) module into each layer of the U-Net based pipeline. This module enables flexible integration of beneficial information from low-interference channels during the processing of high-interference channels, thereby enhancing the denoising capability for the degraded channels. The detailed network structure is illustrated in Fig. \ref{fig:MSA-Net}.

The total parameter count of the proposed MSA-Net is 4.89 MB, and the FLOPs is 109.26GMac, ensuring a good balance between performance and efficiency.

\paragraph{Implementation details}
For the model configuration, the finally submitted model adopts a U-Net structure with a depth of $l=4$. Each layer contains $r=2$ residual blocks, and the channel dimensions are set as $c=[32, 64, 128, 256]$(MSA-Net). To achieve better performance under limited parameters, this model is distilled from a larger teacher network with $r=4$ residual blocks per layer and channel dimensions $c=[64, 128, 256, 512]$ (MSA-Net-D). In terms of training setup, we strictly use only the datasets and loss functions provided in the challenge, without any modifications beyond the proposed network structure. The pre-processing of the dataset follows the standard pipeline specified by the challenge, with no additional custom operations.

\subsection{A Lightweight Multi-Scale Convolutional Attention Network for RAW Image Denoising}
\label{sec:chaos}

\begin{center}

\vspace{2mm}
\noindent\emph{\textbf{Chaos Tamers}}
\vspace{2mm}

\noindent\emph{Shihao Zhou, Sen Yang, Congcong Sun\\Wentao Gu, Jufeng Yang}

\vspace{2mm}

\noindent\emph{CvLab, College of Computer Science, Nankai University}

\vspace{2mm}

\noindent{\emph{Contact: \url{zhoushihao96@mail.nankai.edu.cn}}}

\end{center}

We present a refined U-Net architecture that integrates multi-scale feature extraction with lightweight convolutional attention for real-world RAW image denoising. The encoder progressively captures rich representations through stacked convolution and downsampling layers, while the decoder restores spatial detail via transposed convolutions, enhanced by skip connections that fuse low-level features from the encoder. To better capture both spatial and channel dependencies, we embed convolution-based multi-head attention modules in the decoding path, with learnable scaling factors that adaptively regulate their impact. A bias-free LayerNorm is used throughout to improve numerical stability and generalization. Operating within the challenge’s efficiency constraints, our model achieves 67.36 GMac of computational cost and just 8.13 M parameters, striking an effective balance between denoising performance and inference speed. Finally, our model achieved a PSNR of 42.011 and an SSIM of 0.9639 on the validation set.

\paragraph{Method description}

We propose a hybrid architecture that integrates convolutional neural networks (CNNs) with Transformer-style attention mechanisms for RAW image denoising. The backbone follows a U-Net–style encoder–decoder design, using stacked convolutional layers for multi-scale feature extraction and reconstruction. In the decoder, multi-head attention modules are introduced to capture long-range dependencies and enhance global feature modeling, while LayerNorm ensures stable training. Inspired by U-Net’s multi-scale feature fusion and incorporating elements from Vision Transformers (ViT) and attention-augmented networks, our model achieves a balance between preserving fine local details and capturing global context.

We strictly follow the competition rules, using only the official datasets provided by the organizers. Data preprocessing is handled by SynthTrainDataset, which randomly crops 512×512 patches (8 per image) from clean–noisy pairs, simulates realistic noise based on camera configuration, ISO, and digital gain settings, clips pixel values to valid ranges, normalizes them to [0, 1], and adjusts dimensions to match the model’s input requirements.

Training is performed on a combination of synthetic RAW noise data and the official challenge dataset. We adopt the AdamW optimizer, CosineAnnealingLR scheduler, and L1 loss, combined with mixed precision (fp16) and distributed training via the Accelerate library. The network meets the challenge constraints of fewer than 15M parameters and 150 GMacs while delivering high-quality denoising results.

For inference, we employ a dual-model, multi-strategy pipeline: a “sharp” model incorporating local attention (TLC) and a “faithful” model preserving global attention. Each RAW input undergoes 8 self-ensemble inferences using rotation and flip augmentations, processed independently by both models. Their outputs are fused using predefined weights, producing denoised results that balance sharpness and naturalness. Final outputs are saved in RGGB .npy format for submission.

\begin{figure}[t]
    \centering
    \includegraphics[width=\linewidth]{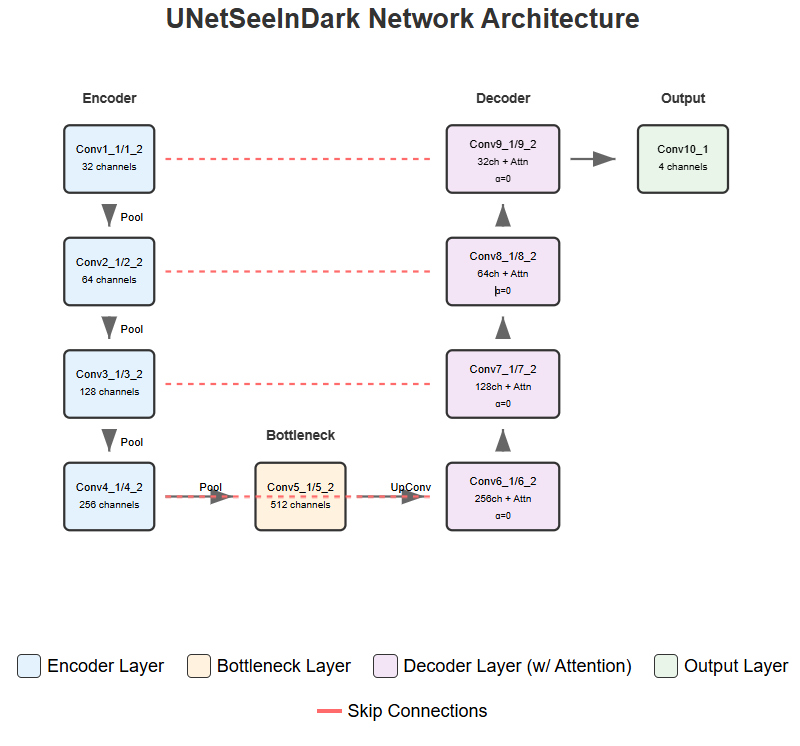}
    \caption{A Lightweight Multi-Scale Convolutional Attention Network for RAW Image Denoising Network.}
    \label{fig:msunet}
\end{figure}

\paragraph{Implementation details}
Our entire pipeline is implemented using the PyTorch framework. We use the AdamW optimizer with an initial learning rate of 2e-4, along with a CosineAnnealingLR scheduler. The training dataset is based on the Sony low-light raw image dataset, and we generate synthetic noisy inputs using the SynthTrainDataset. We train on ISO levels [800, 1600, 3200] with digital gain (dgain) randomly sampled in the range [10, 200]. Each image is randomly cropped into 512×512 patches, and pre-processed with brightness clipping and normalization.

Training is conducted from scratch (no pretraining) using 2 × NVIDIA RTX 4090 GPUs (48GB each) with distributed training via the accelerate library. We use a batch size of 1, training for 500 epochs, with the total training time being approximately 28 hours. Several advanced strategies are employed to improve performance and robustness:
(1) TLC (Local Attention Mechanism): During inference, global attention layers are replaced by localized sliding-window attention to reduce memory consumption and improve detail preservation.

(2) 8x Self-Ensemble Inference: The input undergoes 8 geometric transformations; outputs are inverse-transformed and averaged to enhance robustness.

(3) Dual Model Blending: We combine results from a "sharp" model (with TLC) and a "faithful" baseline model (without TLC) via weighted averaging.

Raw image preprocessing includes black level subtraction, dark shading correction, digital gain application, ROI cropping, RGGB packing, and normalization to the [0,1] range. 

\section*{Acknowledgments}
This work was partially supported by the Alexander von Humboldt Foundation. We thank the AIM 2025 sponsors: AI Witchlabs and University of W\"urzburg (Computer Vision Lab).

{
    \small
    \bibliographystyle{ieeenat_fullname}
    \bibliography{main.bib}
}

\end{document}